# Bayesian Optimal Active Search and Surveying

**Roman Garnett**　　　　　　　　　　　　　　　　　　　　　　　　　　RGARNETT@CS.CMU.EDU
**Yamuna Krishnamurthy**　　　　　　　　　　　　　　　　　　　　　　YKRISHNA@ANDREW.CMU.EDU
**Xuehan Xiong**　　　　　　　　　　　　　　　　　　　　　　　　　　XXIONG@ANDREW.CMU.EDU
**Jeff Schneider**　　　　　　　　　　　　　　　　　　　　　　　　　　JSCHNEIDE@CS.CMU.EDU
Carnegie Mellon University, 5000 Forbes Avenue, Pittsburgh, PA 15213, United States

**Richard Mann**　　　　　　　　　　　　　　　　　　　　　　　　　　RMANN@MATH.UU.SE
Uppsala Universitet, Uppsala 751 06, Sweden

## Abstract

We consider two active binary-classification problems with atypical objectives. In the first, *active search*, our goal is to actively uncover as many members of a given class as possible. In the second, *active surveying*, our goal is to actively query points to ultimately predict the proportion of a given class. Numerous real-world problems can be framed in these terms, and in either case typical model-based concerns such as generalization error are only of secondary importance.

We approach these problems via Bayesian decision theory; after choosing natural utility functions, we derive the optimal policies. We provide three contributions. In addition to introducing the active surveying problem, we extend previous work on active search in two ways. First, we prove a novel theoretical result, that less-myopic approximations to the optimal policy can outperform more-myopic approximations by any arbitrary degree. We then derive bounds that for certain models allow us to reduce (in practice dramatically) the exponential search space required by a naïve implementation of the optimal policy, enabling further lookahead while still ensuring that optimal decisions are always made.

## 1. Introduction

In many real-world classification scenarios, it can be much easier to collect input data than to observe associated labels, which could require relatively expensive human action. For this reason, considerable research in semi-supervised and active learning has considered how to construct models exploiting unlabeled data and also how to intelligently request unknown labels to achieve a given goal as cheaply as possible.

The bulk of active classification research has considered obtaining labels to maximize some measure of predictive power or model accuracy. Here, we consider two distinctly different problems. In the first, which we call *active search*, the members of one particular class are deemed important and are to be located as quickly as possible. Many real-world problems are of this form; fraud detection, drug discovery, and product recommendation are just a few examples. In the second task, which we call *active surveying*, we seek to determine the portion of a dataset belonging to a particular class. Targeted opinion polling is an important and natural real-world problem of this type.

Typical model-based active classification strategies are not appropriate for either of these problems. The consequence of catching a fraudster, discovering a new cancer drug, or selling a product can be measured in monetary terms. Learning an accurate model, on the other hand, is only useful if it can help us locate more items. Indeed we could make observations that give very high performance on either task and nonetheless produce a model that is uncertain or even completely inaccurate on large swathes of the domain.

Rather than proposing heuristics to adapt typical active learning algorithms to these problems, we will instead begin "from the beginning" and analyze these problems using Bayesian decision theory. We will first define natural utility functions for each problem and then derive the optimal policies.

The active search problem has been previously described (Garnett et al., 2011); here we extend that preliminary work with two contributions. First, we will prove that a myopic approximation to the optimal active search policy can perform arbitrary worse than an even slightly less-myopic approximation. Our second contribution is more practical. In the general case, the optimal search policy requires time that grows exponentially in the number of unlabeled points. Here we show how for a certain class of classifiers (including $k$-NN), we may identify and discard points that cannot possibly be optimal with trivial extra computation. In practice, this can increase the efficiency of the algorithm by orders of magnitude and allow us to use policies that might not otherwise be possible.

The rest of this paper is arranged as follows. In Sections 2 and 3, we formally describe the problems at hand. In Section 4, we provide and discuss the Bayesian optimal policies for these problems. We proceed by proving a result about the potential benefit of using the optimal active search policy with increasingly long horizons. In Section 6, we discuss a branch-and-bound technique to limit the search space required for the optimal search policy. Finally, we evaluate our methods empirically.

## 2. Problem Definition

Suppose we have a finite set of elements $\mathcal{X} \triangleq \{x_i\}$ and an identified subset $\mathcal{R} \subset \mathcal{X}$, the members of which we will call *targets*. We consider the following problem. Suppose we do not know which members of $\mathcal{X}$ belong to $\mathcal{R}$ *a priori,* but can successively request binary observations $y \triangleq \chi(x \in \mathcal{X})$, for an unlabeled element $x \in \mathcal{X}$. We wish to actively select a sequence of queries to maximize a given utility function.

For the active search problem, we define the utility of a set of observations $\mathcal{D} \triangleq \{(x_i, y_i)\}$ to be the number of targets found:

$$u(\mathcal{D}) \triangleq \sum y_i.$$

This simple expression naturally captures the spirit of the problem as defined above. For the active surveying problem, we define the utility of a set of observations to be the variance in our induced probability distribution over the cardinality of $\mathcal{R}$:

$$u(\mathcal{D}) \triangleq -\operatorname{var}[\operatorname{card} \mathcal{R} \mid \mathcal{D}].$$

Again this expression encapsulates the goal of surveying: polls with smaller margins of error are to be preferred.

## 3. Related Work

Active learning is a mature field with a large associated body of literature (Settles, 2010). In the active binary-classification problem, the chosen objective is usually related to properties of the associated probabilistic model. Examples include generalization error (Zhu et al., 2003) and optimality criteria related to the Fisher information, such as A-optimality (Schein & Ungar, 2007). One of the simplest active learning techniques for binary classification is *uncertainty sampling* (Lewis & Gale, 1994), which successively requests the label for the point $x^*$ with the greatest posterior variance: $x^* \triangleq \arg\min_x |\Pr(y = 1 \mid x, \mathcal{D}) - 1/2|$.

Both objectives considered in this paper are unusual in an active-learning context as far as the authors know. A problem similar to active search that has been considered is the active discovery of previously unseen classes (He & Carbonell, 2008). Weitzman (1979) considers an active search problem where there is no dependence between outcomes and derives the optimal policy. The problem as defined there can also be seen as a Bayesian multi-armed bandit, and the optimal policy can also be recovered via a Gittins index (Gittins et al., 2011). Here we consider the case where the "arms" of the bandits are correlated, which is a so-called *restless* bandit problem.

There is a long history of statistical research investigating the selection of respondents when conducting a survey. A particular focus of such research is identifying and correcting *selection bias*, where certain people are more likely to be selected for a survey than others (Berger, 2005). Here we take a completely different approach: we actively and intentionally bias our selection of points to query, choosing those that we believe will increase our understanding of the class proportion as much as possible. In the context of polling a social network, we can reasonably expect that opinions are in some way correlated under a notion of "similarity" or "closeness" in the network. For this reason, polling people with many connections throughout the network might be more fruitful than polling people who are relatively isolated. We embrace and leverage this notion of correlated opinions and influence in our design.

## 4. The Optimal Bayesian Policy

As mentioned previously, our approach to the active search and surveying problems will be motivated by Bayesian decision theory. This will require selecting a classification model that provides the posterior probability of a point $x$ belonging to $\mathcal{R}$ conditioned on previously observed data $\mathcal{D}$, $\Pr(y = 1 \mid x, \mathcal{D})$. We will assume that this model is given *a priori*; the decision theoretic analysis does not depend on its nature.

Without loss of generality, we will assume that at the onset we will be allowed a fixed number of queries $t$.

In applications where the cost of obtaining a label is high, it is the total cost of the queries that limits their number, rather than the quantity of unlabeled points.

We now derive the policy for deciding the locations of our queries, which will entail successively calculating the expected utility of each of the remaining unlabeled points then observing the label for the point that with maximal expected utility. At time $i$, then, we will observe the label for the point

$$x_i^* \triangleq \underset{x_i \in \mathcal{X} \setminus \mathcal{D}_{i-1}}{\arg \max} \; \mathbb{E}\big[u(\mathcal{D}_t) \mid x_i, \mathcal{D}_{i-1}\big].$$

We begin by considering the case when we are allowed to make exactly one more query and will then address the general case. Suppose that we have already made $t-1$ observations $\mathcal{D}_{t-1}$. To select our final observation, we calculate the expected utility of a candidate point $x_t$, marginalizing out the unknown value of $y_t$.

For active search, the expected utility is

$$\mathbb{E}\big[u(\mathcal{D}_t) \mid x_t, \mathcal{D}_{t-1}\big] = \sum_y u(\mathcal{D}_t) \Pr(y_t = y \mid x_t, \mathcal{D}_{t-1})$$
$$= u(\mathcal{D}_{t-1}) + \Pr(y_t = 1 \mid x_t, \mathcal{D}_{t-1}).$$

Because $u(\mathcal{D}_{t-1})$ does not depend on $x_t$, the optimal decision $x_t^*$ is therefore the point with the largest posterior probability of being a target. This makes intuitive sense: with only one evaluation remaining, there is no possible benefit to explore, and we might as well make a purely greedy last try.

For active surveying, the expected utility is

$$\sum_y u(\mathcal{D}_t) \Pr(y_t = y \mid x_t, \mathcal{D}_{t-1})$$
$$= \mathbb{E}_{y_t}\Big[-\mathrm{var}[\mathrm{card}\,\mathcal{R} \mid \mathcal{D}_t] \mid x_t, \mathcal{D}_{t-1}\Big].$$

The optimal decision $x_t^*$ is therefore the point with the smallest expected variance of $p(\mathrm{card}\,\mathcal{R} \mid \mathcal{D}_t)$. This is a bit more opaque than the active search expression above, but is still intuitively reasonable.

Given the optimal policy for selecting $x_t$, we now consider the problem of choosing the location of the second-to-last point $x_{t-1}$. When making our decision in this case (as well as with any other $x_i$ with $i < t$), the problem becomes more difficult because we must now contemplate the possible consequences of our choices and how they will impact our future decisions. The mechanical manifestation of this remark is that during the calculation of the expected utility for the two-step lookahead case, we must integrate out the unknown location of the final observation $x_t$, as well as its label:

$$\mathbb{E}\big[u(\mathcal{D}_t) \mid x_{t-1}, \mathcal{D}_{t-2}\big] =$$
$$\iiint u(\mathcal{D}_t) \Pr(y_{t-1} \mid x_{t-1}, \mathcal{D}_{t-2}) p(x_t \mid \mathcal{D}_{t-1}) \cdots$$
$$\cdots \Pr(y_t \mid x_t, \mathcal{D}_{t-1}) \, \mathrm{d}y_{t-1} \, \mathrm{d}x_t \, \mathrm{d}y_t. \quad (1)$$

Note, however, that the integral over $x_t$ can be evaluated trivially because $p(x_t \mid \mathcal{D}_{t-1})$ is simply $\delta(x_t - x_t^*)$,[1] where $\delta$ is the Dirac delta function—that is, given the value of $y_{t-1}$, the location of the last choice $x_t$ is deterministic and known from our discussion above.

To evaluate the two-step expected utility at a point $x_{t-1}$, we therefore sample over the unknown value $y_{t-1} \in \{0, 1\}$; for each possible value of $y_{t-1}$, we find the optimal last observation $x_t^*$ given that fictitious observation as described above. Note that sampling over $y_t^*$ is not required in the search case.

We may repeat the procedure described above recursively to calculate the expected $\ell$-step lookahead utility of choosing a point for any $\ell \leq t$, allowing us to operate on any horizon. We note that some authors would equivalently discuss the preceding analysis in terms of Bellman's equation and Markov decision processes (MDPs); our choice of presentation is purely stylistic.[2]

As noted in (Garnett et al., 2011), the optimal policies for both of these problems in general requires running time $\mathcal{O}\big((2\,\mathrm{card}\,\mathcal{X})^\ell\big)$. For lookahead more than a few steps into the future, this procedure can become daunting due to the sampling required. This is a common issue in sequential Bayesian decision problems. One typical way to address this problem is to approximate exact inference by shortening our horizon (Jones et al., 1998; Osborne et al., 2009). For timestep $t - m$ with $m > \ell$, we myopically pretend that there are only $\ell$ observations remaining and choose $x_{t-m}$ by maximizing the $\ell$-step lookahead expected utility. We will address this issue further in Section 6 and show how in some reasonable cases we can restrict the exponential search space required to find the $\ell$-step optimal decision.

## 5. Potential Gain from Looking Ahead

In this section we will discuss the behavior of the $\ell$-step optimal search policy versus the $m$-step policy, for $\ell < m$. Let us first consider the behavior of the two-step policy versus the simple greedy one-step policy. Notice that the two-step policy allows us to make decisions that do not maximize the posterior probability of observing a target at the current step. Instead, we might choose to explore a region where the probability of immediate reward is lower, but where there is a chance of discovering more targets overall during the next two evaluations. We will give a very simple example that demonstrates the effect of this tradeoff in the active search case. Figure 1 shows a three-point space. The two connected points have the same label; they are known to either both be targets or both be nontargets, with marginal probability of their being targets $\varepsilon$. The point on the right is independent of the others and has probability of being a target $\delta > \varepsilon$. Consider being allowed two label queries with the goal of locating as many targets as possible. We may calculate the expected performance of the one- and two-step policies directly. The one-step policy will always choose the right point first and then will be compelled to choose either of the left points, with expected final utility $\varepsilon + \delta$. The expected two-step utility of the left points is each $2\varepsilon + (1 - \varepsilon)\delta$, and the expected two-step utility of the right point is $\varepsilon + \delta$. The difference in two-step expected utility between either left point and the right point is $\varepsilon(1 - \delta) > 0$; therefore one of the left points will always be chosen, and the two-step policy will outperform the one-step greedy policy on average for any value of $\delta$.

---

[1] Note that $x_t^*$ depends on the unknown value of $y_{t-1}$.

[2] In the MDP form, the potential for computational savings via dynamic programming is more apparent; however, the state space is still exponential in $t$.

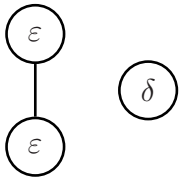

Figure 1: The simple probability model used in the discussion comparing the typical behavior of the optimal policy versus a greedy policy. The points on the left are known to have the same label, and the marginal probability of that label is $\Pr(y = 1) \triangleq \varepsilon$. The label of the solitary point on the right is independent of the others with probability $\Pr(y = 1) \triangleq \delta > \varepsilon$.

This example demonstrates the sort of nontrivial decisions that the optimal policy can make—it can be better to explore a region where labels are expected to be highly correlated, even when the probability of being "lucky" and finding many targets is much smaller than the current most likely single point. A welcome side effect of this behavior is that when such a decision is made, we learn about the labels of the points in the chosen region even if we do not observe a target. Such evaluations can therefore be advantageous by our having improved the overall quality of our probabilistic model, despite this goal not having been specified at any step during our derivation of the optimal policy.

We can extend the ideas in the above example to prove that, in the case of active search, increasing our horizon can always improve performance by any arbitrary degree. Let $\mathcal{P} \triangleq (\Omega, 2^\Omega, \Pr)$ be a (discrete) probability space on $\Omega$. Given $\mathcal{P}$, we will denote the expected utility of the $\ell$-step-lookahead policy after $t$ evaluations with $\mathbb{E}_{\mathcal{D}}\big[u(\mathcal{D}) \mid \ell, t, \mathcal{P}\big]$. We may prove the following.

**Theorem 1.** *Let $\ell, m \in \mathbb{N}^+, \ell < m$. For any $q > 0$, there exists a $\mathcal{P}$ and $t$ such that*

$$\frac{\mathbb{E}_{\mathcal{D}}\big[u(\mathcal{D}) \mid m, t, \mathcal{P}\big]}{\mathbb{E}_{\mathcal{D}}\big[u(\mathcal{D}) \mid \ell, t, \mathcal{P}\big]} > q;$$

*that is, the $m$-step active-search policy can outperform the $\ell$-step policy by any arbitrary degree.*

*Proof.* As in the simple example above, the key to the argument is that $\ell$-step lookahead cannot differentiate between a "clump" of correlated points of size $\ell$ and one of size greater than $\ell$. To formalize this concept, define a $(k, \varepsilon)$-*clump*, denoted $\textup{\textcircled{$k$}}_\varepsilon$, to be a collection of $k$ discrete points that all have the same label, with marginal probability of being all targets $\varepsilon$.

Consider applying the $m$-step lookahead policy for querying $t$ labels on the space

$$\mathcal{P} \triangleq \bigsqcup_{i=1}^{t} \textup{\textcircled{$t$}}_\varepsilon.$$

The policy is easy to analyze in this case. After no evaluations, every point in the domain has the same expected $m$-step utility by symmetry. After observing that point, either a clump of targets will have been discovered (with probability $\varepsilon$), or a clump of nontargets. In the former case, the remaining $t - 1$ evaluations will all be spent querying the remaining points in the selected clump, because they will all have maximal expected utility for all horizons. In the latter case, a point in another unobserved clump will be chosen, and the response to the outcome will be the same. Given this, we may calculate:

$$\mathbb{E}_{\mathcal{D}}\big[u(\mathcal{D}) \mid m, t, \mathcal{P}\big] = \sum_{i=1}^{t} \varepsilon(1-\varepsilon)^{i-1}(t - i + 1). \quad (2)$$

We now augment $\mathcal{P}$ with $t$ copies of $\textup{\textcircled{$\ell$}}_\delta$, with $\delta > \varepsilon$:

$$\mathcal{P} \triangleq \left(\bigsqcup_{i=1}^{t} \textup{\textcircled{$t$}}_\varepsilon\right) \cup \left(\bigsqcup_{i=1}^{t} \textup{\textcircled{$\ell$}}_\delta\right).$$

It is trivial to show that the $\ell$-step utility of a point in a $\textup{\textcircled{$\ell$}}_\delta$ is greater than the $\ell$-step utility of a point in a $\textcircled{t}_\varepsilon$. The form for each is of the same form as (2), and their difference may be calculated directly; it is

$$(1-\varepsilon)^\ell - (1-\delta)^\ell + \delta^{-1}\big((1-\delta)^\ell - 1\big) - \varepsilon^{-1}\big((1-\varepsilon)^\ell - 1\big) > 0.$$

Despite the fact that the $t$-clump has more potential targets, the $\ell$-step lookahead policy greedily chooses the more immediately fruitful $\ell$-clump.

With this, we may find an upper bound for $\mathbb{E}_\mathcal{D}\big[u(\mathcal{D}) \mid \ell, t, \mathcal{P}\big]$. Consider build a string $S$ (initially empty) as follows. Sample $r$ from $\mathcal{U}(0,1)$. If $r < \delta$, append $\ell$ 1s to $S$; otherwise, append a 0. Repeat a $k$ times. At termination, the probability of a character in $S$ being a 1 does not depend on $k$; it is $\delta\ell/\delta(\ell-1)+1$. We can simulate the $\ell$-step method by stopping when length$(S) \geq t$ and taking the first $t$ characters. When we stop, the expected number of 1s in the first $t$ can obviously not be greater than the expected number in $S$, which has length at most $(t+\ell-1)$:

$$\mathbb{E}_\mathcal{D}\big[u(\mathcal{D}) \mid \ell, t, \mathcal{P}\big] < \frac{(t+\ell)\delta\ell}{\delta(\ell-1)+1}. \tag{3}$$

The final component of the proof is showing that even if $\varepsilon < \delta$, the $m$-step lookahead expected utility of a point in one of the $t$-clumps is greater than the $m$-step lookahead expected utility of a point in one of the $\ell$-clumps. One can in fact prove the following, which generalizes the situation in Figure 1 (consider as $\epsilon \to \delta^-$).

**Lemma 1.** *Let $\ell, m, k \in \mathbb{N}^+$, $\ell < m \leq k$, and $\delta \in (0,1)$ be given. Then there is a $\varepsilon < \delta$ such that the expected $m$-step utility of a point in an unobserved $\textcircled{k}_\varepsilon$ is greater than that of a point in an unobserved $\textcircled{\ell}_\delta$.*

Set $\varepsilon < \delta$ such that the $m$-step policy selects the $\textcircled{t}_\varepsilon$ clumps. Notice that the $m$-step policy will behave identically as before, except that it can now switch to the $\textcircled{\ell}_\delta$ clumps with fewer than $m$ evaluations remaining in "unlucky" cases. The right-hand side of (2) therefore still serves as a lower bound on its performance in this new space.

Finally, combining the lower bound in (2) and the upper bound in (3), we have

$$\frac{\mathbb{E}_\mathcal{D}\big[u(\mathcal{D}) \mid m, t, \mathcal{P}\big]}{\mathbb{E}_\mathcal{D}\big[u(\mathcal{D}) \mid \ell, t, \mathcal{P}\big]} > \frac{\big((1-\varepsilon)^{t+1} + \varepsilon t + \varepsilon - 1)\big)\big(\delta(\ell-1)+1\big)}{\varepsilon\delta\ell(t+\ell)},$$

which may be made arbitrarily large by taking small enough $\delta$ and large enough $t$. $\square$

## 6. Bounding the Active Search Space

We will now discuss how we may, in certain situations, reduce the $\mathcal{O}\big((2\operatorname{card}\mathcal{X})^\ell\big)$ search space required by the $\ell$-step optimal active search policy. Our approach will entail a "branch and bound"–style strategy, where we will leverage relatively inexpensive-to-calculate inequalities to prune suboptimal branches of the search space from consideration. This will require establishing two inequalities. First, we find a lower bound on the maximal $\ell$-step active search expected utility among the unlabeled points. Next we find an upper bound on the $\ell$-step expected utility of a given unlabeled point, as a function of its current probability. Combining these bounds together will ultimately provide us with a threshold $\theta$ such that any point $x$ with $\Pr(y = 1 \mid x, \mathcal{D}) < \theta$ cannot possibly be the optimal $\ell$-step action. In the below we will assume we start at timestep 1 and progress to timestep $\ell$, beginning with an arbitrary starting set $\mathcal{D}_1$.

### 6.1. A lower bound on $\max \mathbb{E}\big[u(\mathcal{D}_\ell) \mid x_1, \mathcal{D}_1\big]$

We will first establish a trivial lower bound on the maximal $\ell$-step expected utility. Let

$$(x', p') \triangleq (\arg)\max_{x \in \mathcal{X}\setminus\mathcal{D}_1} \Pr(y = 1 \mid x, \mathcal{D}_1)$$

be the point with the highest posterior probability of being a target at the first timestep, along with its probability. Let

$$u' \triangleq \mathbb{E}\big[u(\mathcal{D}_\ell) \mid x_1 = x', \mathcal{D}_1\big]. \tag{4}$$

Clearly then $u'$ is a trivial bound on the maximal expected $\ell$-step utility.

### 6.2. An upper bound on $\mathbb{E}\big[u(\mathcal{D}_\ell) \mid x_1, \mathcal{D}_1\big]$

We now find an upper bound on the $\ell$-step expected utility for any arbitrary point $x_1$. Our approach will require the chosen classification model to meet two conditions. First, we must have that conditioning on a new nontarget observation cannot raise the target probability for any unlabeled point. Second, we must be able to bound the maximum target probability among unlabeled points after conditioning on a given number of additional targets; that is, we assume there is a function $p^*(n, \mathcal{D})$ such that

$$p^*(n, \mathcal{D}) \geq \max_{x \in \mathcal{X}\setminus\mathcal{D}} \Pr\big(y = 1 \mid x, \mathcal{D}\cup\mathcal{D}', \textstyle\sum_{y' \in \mathcal{D}'} y' \leq n\big). \tag{5}$$

With the $p^*$ function in hand, we will define a function $u^*(\ell, n, \mathcal{D})$ that represents a bound on the maximum $\ell$-step utility among any unlabeled point after $n$ addi-

Bayesian Optimal Active Search and Surveying

tional target observations. For $\ell = 1$, we define

$$u^*(\ell = 1, n, \mathcal{D}) \triangleq p^*(n, \mathcal{D}).$$

That this bound is valid follows immediately from the analysis of the simple one-step active search case. For $\ell > 1$, we may build $u^*$ recursively:

$$u^*(\ell, n, \mathcal{D}) \triangleq p^*(n, \mathcal{D})\bigl(u^*(\ell - 1, n + 1, \mathcal{D}) + 1\bigr) + \bigl(1 - p^*(n, \mathcal{D})\bigr)u^*(\ell - 1, n, \mathcal{D}).$$

With $u^*$ now defined, for a given point $x$, we have

$$\mathbb{E}\bigl[u(\mathcal{D}_\ell) \mid x_1 = x, \mathcal{D}_1\bigr] \leq$$
$$\Pr(y = 1 \mid x, \mathcal{D})\bigl(u^*(\ell - 1, 1, \mathcal{D}) + 1\bigr) +$$
$$\bigl(1 - \Pr(y = 1 \mid x, \mathcal{D})\bigr)u^*(\ell - 1, 0, \mathcal{D}). \quad (6)$$

Combining (4) and (6), we may eliminate any point such that the right-hand side of (6) is less than $u'$, because it cannot possibly be the optimal action. Of course, we may apply this pruning technique at all depths of the search tree, allowing for deeper suboptimal subtrees to be found and eliminated as well.

## 7. Results

We implemented the optimal active search and surveying policies in MATLAB, as well as uncertainty sampling. Using this implementation, we evaluated the performance of our policies on both synthetic and real data.

In our search experiments, we used a simple $k$-nearest neighbor classifier. Let $\text{NN}(x)$ represent the $k$-nearest neighbors of the point $x$ in $\mathcal{X}$, and let $\text{L-NN}(x)$ represent the subset of $\text{NN}(x)$ for which we currently have label observations. We define

$$\Pr(y = 1 \mid x, \mathcal{D}) \triangleq \frac{\gamma + \sum_{x' \in \text{L-NN}(x)} y'}{1 + \sum_{x' \in \text{L-NN}(x)} 1}. \quad (7)$$

Here the constant $\gamma \in [0, 1]$ serves as a "pseudocount," which smooths the probabilities on points that have few labeled neighbors. In our experiments, we fixed $\gamma \triangleq 1/10$. This model worked well empirically, and we may also easily derive the bound on maximum probabilities in (5) required for pruning the search space as described above. If we consider a point $x$ with current probability

$$\Pr(y = 1 \mid x, \mathcal{D}) \triangleq \frac{\gamma + \alpha}{1 + \beta},$$

then after conditioning on new observations $\mathcal{D}'$ containing at most $n$ more positive observations, we have

$$\Pr\bigl(y = 1 \mid x, \mathcal{D} \cup \mathcal{D}', \textstyle\sum_{y' \in \mathcal{D}'} y' \leq n\bigr) \leq \frac{\gamma + \alpha + n}{1 + \beta + n}.$$

Note that this bound can be trivially modified to allow for arbitrary weight functions to be included in the model; there $n$ could be replaced with $n\bigl(\max_{x' \in \text{NN}(x)} w(x, x')\bigr)$.

### 7.1. Illustrative example

We begin with a simple example problem that illustrates the behavior of the active search and active surveying approaches versus uncertainty sampling.

Let $I \triangleq [0, 1]^2$ be the unit square. We repeated the following experiment 100 times. We selected 250 points uniformly at random from $I$, which formed our input space $\mathcal{X}$. Any point landing within Euclidean distance $1/4$ of any of the points $(0, 0)$, $(0, 1)$, $(1, 0)$, $(1, 1)$ or $(1/2, 1/2)$ (the four corners and the center point) formed the set of targets $\mathcal{R}$. We picked one point uniformly at random from $\mathcal{R}$ and added it and its label to a training set. We then used the one-step optimal active search policy, the one-step optimal active surveying policy, and uncertainty sampling to select ten more points.

Figure 2 shows kernel density estimates of the points selected by the algorithms across all experiments. The difference in behavior is immediate. Uncertainty sampling strongly focuses on the corners, where variance is typically the highest, the search policy strongly focuses on the learned locations of the targets, including the center, and the surveying policy strongly avoids the corners, which, despite having high variance, are not terribly informative about the space overall.

### 7.2. CiteSeer[x] data

For our next experiment, we created a graph from a subset of the CiteSeer[x] citation network. Papers in the database were grouped based on their venue of publication (after extensive data cleaning), and papers from the 48 venues with the most associated publications were retained. The graph was defined by having these papers as its nodes (38 079 in total) and undirected citation relations as its edges. We designated all papers appearing in NIPS proceedings (2 198 in total, 5.2% of the dataset) as targets. Differentiating these papers is difficult; many highly related venues are also prevalent.

For our results presented here, we computed what Fouss et al. (2007) calls "graph principal component analysis," which is equivalent to performing principal component analysis on card $V$ vectors (one corresponding to each node) embedded in $\mathbb{R}^{(\text{card }V)-1}$ that are separated by a graph metric called commute time.[3] The first 20 graph principal components formed our set $\mathcal{X}$, and our model was as in (7) with $k \triangleq 50$.

---

[3] The *commute time* between nodes $v$ and $v'$ is the expected time a simple random walk beginning at $v$ takes to hit $v'$ and return.

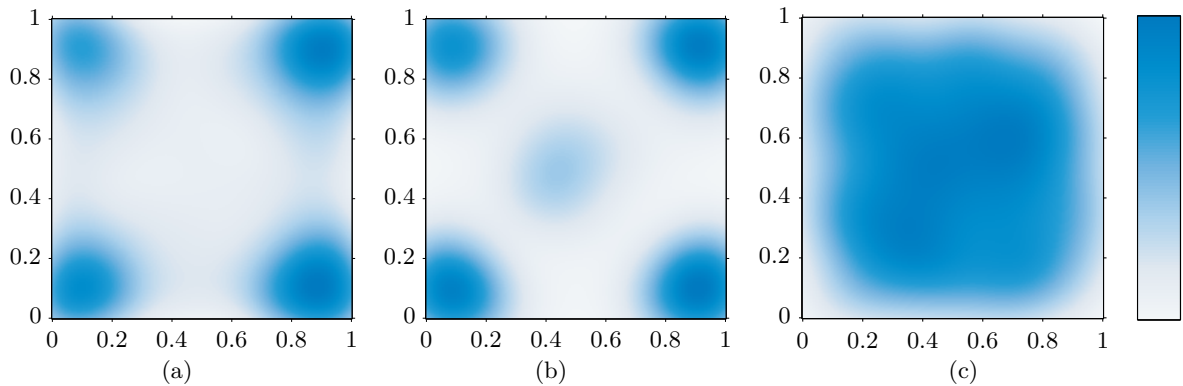

Figure 2: Kernel density estimates of the distribution of points chosen by (a) uncertainty sampling, (b), one-step optimal active search, and (c), one-step optimal active surveying for the simple two-dimensional demonstrative example. Darker blue indicates more probability mass.

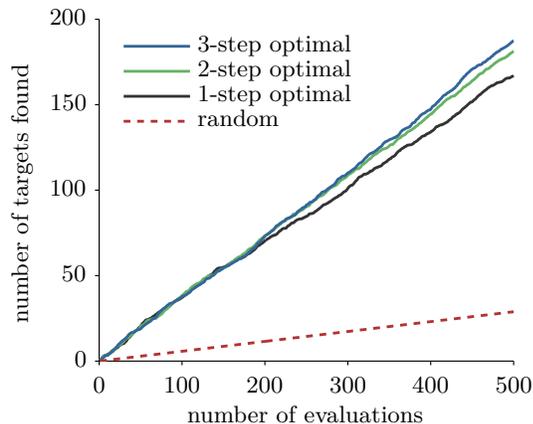

Figure 3: Cumulative number of targets found during 1 000 steps of several active querying schemes on the CiteSeer[x] data. The dashed red line shows the expected performance of random sampling.

Again, we selected a single point at random from $\mathcal{R}$ to form an initial training set, then ran 500 steps of the one-, two-, and three-step active search policies with the goal of finding as many NIPS papers as possible. This experiment was repeated ten times.

On average, the one-step algorithm found 167 NIPS papers; the two-step algorithm found 180, and the three-step algorithm found 187. Random search would be expected to find only 29 papers given the same number of evaluations. Figure 3 shows the cumulative number of targets found by each of the methods. The three-step lookahead procedure was able to find 8.5% of the targets after scanning only 1.3% of the data, 6.5 times better than expected by random search.

To test active surveying, we selected 75 evaluations (starting again with a single training point from $\mathcal{R}$) using three different approaches: random search, uncertainty sampling, and the one-step optimal active-surveying policy. To estimate the class proportion, we subsampled the remaining unlabeled points, selecting 5% each time, and averaged the inferred means and variances of $p(\text{card}\,\mathcal{R} \mid \mathcal{D})$ from five such samples.

After each evaluation of each method, we estimated the mean and variance of $\text{card}\,\mathcal{R}$ given the training data collected thus far, as described above. We evaluated each method's performance by approximating the posterior over the class proportion $\text{card}\,\mathcal{R}/\text{card}\,\mathcal{X}$ with a beta distribution whose parameters $(\alpha, \beta)$ were selected via moment-matching to the mean and variance of the induced posterior distribution over this quantity. We then computed the likelihood of the true unknown class proportion under this beta distribution. Figure 4 shows the progression of these likelihoods for each method over the course of the experiment. After a period of time where all methods have similar performance, the optimal policy begins to significantly outperform the other two methods, which behave nearly identically. There is clear utility to our active-surveying approach, even when the number of samples taken is very small.

**7.3. The effect of search-space pruning**

Finally, we measured the effect of our branch-and-bound method described in Section 6. With the same data and experimental setup as in the CiteSeer[x] experiment, we measured the time required by one iteration of the optimal $\ell$-step lookahead search policy, for $2 \leq \ell \leq 4$, both with and without the advantage of our pruning method. These times were measured given 100 random starting configurations, chosen as before.

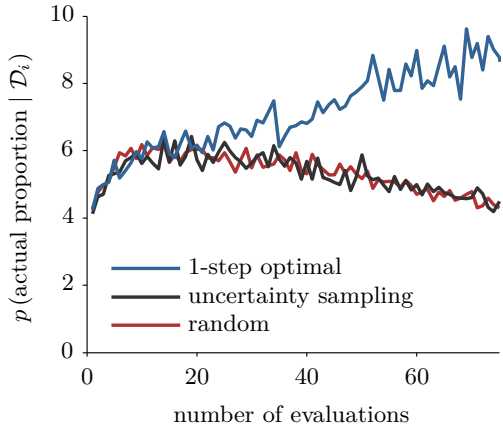

Figure 4: Likelihood of the true class proportion under the moment-matched beta distribution fit to the predictions made after each of the first 75 evaluations made by each method in the CiteSeer[x] experiment.

Table 1: The average time (in seconds) taken for one iteration of the $\ell$-step lookahead optimal search policy on the CiteSeer[x] data, for $1 \leq \ell \leq 4$. Some times are approximate. For reference, the one-step policy took an average of $2.24 \times 10^{-3}$ s per iteration.

|  | $\ell = 2$ | $\ell = 3$ | $\ell = 4$ |
| --- | --- | --- | --- |
| pruning | 0.228 s | 15.0 s | 745 s |
| no pruning | 166 s | ≈146 days | ≈30 500 years |
| speedup | 731 | $8.42 \times 10^5$ | $1.29 \times 10^9$ |

The results are summarized in Table 1. The effect of our pruning strategy in this case is dramatic, enabling us to extend our search horizon far beyond what the realm of possibility would have been otherwise.

## 8. Conclusion

We have presented the Bayesian optimal policy to two atypical active-learning problems related to binary classification, which we call active search and active surveying. The former focuses on actively seeking out members of a set of identified targets as quickly as possible, and the latter focuses on predicting the portion of the dataset belonging to an identified class. Our approach was to define sensible utility functions for these problems and then to derive the optimal Bayesian policy for each of them. The optimal policy for each takes the same form, but in practice the behavior of each can be dramatically different due to the sharply contrasting underlying utility functions.

In addition to introducing the active surveying problem, we have extended previous preliminary work on active search in two ways. We first proved a theoretical result showing that the potential advantage of farther lookahead horizons is unbounded. We then presented a branch-and-bound method for pruning the exponential search space required for active search in certain cases, which we showed can improve the computational performance of the optimal policy by orders of magnitude.